\documentclass[runningheads]{llncs}
\usepackage[T1]{fontenc}
\usepackage{graphicx}
\usepackage{booktabs}
\usepackage{multirow}
\usepackage{amsmath}
\usepackage{amssymb,amsfonts}
\begin{document}
\title{SLM-Conditioned Hierarchical Relation Routing for Labeled Property Graph Learning}
\titlerunning{SLM-Conditioned Hierarchical Relation Routing for LPG Learning}
%
\author{
Michal Podstawski
}
    
\authorrunning{M. Podstawski}

\institute{
NASK National Research Institute, Warsaw, Poland\\
\url{https://nask.pl/}
\\
\email{michal.podstawski@nask.pl}
}
\maketitle              
\begin{abstract}
Labeled property graphs combine relational structure with heterogeneous textual and categorical properties attached to both nodes and relationships. Conventional graph neural networks typically represent these properties as static feature vectors, limiting their ability to determine which semantic evidence should influence message propagation for a particular prediction target. We propose SLM-Conditioned Hierarchical Relation Routing, an architecture that integrates a small language model directly into graph message selection. A topology GNN provides a stable structural representation and prediction anchor. For each target node, incident messages combine the neighbor’s structural state, node-property encoding, relationship-property encoding, and relationship type. A parameter-efficient SLM processes structured graph soft tokens and produces a target-conditioned routing query. This query first selects relevant messages within each relationship type and subsequently routes information across relation-level summaries. The resulting representation provides a bounded residual update to the topology anchor, preserving structural evidence while allowing contextual semantic information to modify the prediction. The architecture supports interpretable analysis at both the neighbor and relationship-type levels and provides a general mechanism for integrating language-derived semantics into property-rich graph learning.

\end{abstract}
\section{Introduction}

Labeled property graphs (LPGs) represent entities and relationships together
with labels and heterogeneous properties. Unlike conventional attributed graphs,
LPGs may associate rich semantic information with both nodes and relationships:
a relationship can communicate not only that two entities are connected, but also
the context, category, time, amount, or description of that connection. This
representation underlies many operational graph databases, with applications in
healthcare, financial analysis, recommendation, cybersecurity, and network
management.

Graph neural networks (GNNs) are an effective mechanism for learning from
relational structure, but common architectures either ignore textual properties
or encode them once and treat the resulting vectors as static features. Such
integration cannot determine whether a particular property, neighboring entity,
or relationship type is relevant to the current prediction target. This is
especially limiting in heterogeneous LPGs, where a single node may participate
in many semantically different relationships and only a few of them carry the
evidence needed for a given prediction.

Language models offer contextual representations of textual and structured
properties, yet existing ways of combining them with graphs sit at two
extremes. Used only as feature encoders, they separate semantic interpretation
from graph propagation, so the interpretation cannot adapt to the target.
Used to read entire neighborhoods rendered as text, they discard explicit graph
structure and quickly exceed the context window. Neither lets language-derived
context control \emph{which} relational evidence flows during message passing.
A useful integration should preserve graph topology while allowing semantics to
modulate propagation conditionally on the target.

We introduce \emph{SLM-Conditioned Hierarchical Relation Routing}, an
architecture in which a small language model (SLM) directly conditions GNN
message selection. A topology GNN first produces a stable structural
representation and prediction anchor. For each target, incident messages combine
structural neighbor states, semantic node and relationship encodings, and
relationship-type information. A parameter-efficient SLM derives a
target-conditioned query that routes these messages in two stages - first within
individual relationship types, then across relation-level summaries - and the
routed evidence produces a bounded residual update to the anchor. Because the
correction is bounded and starts from the anchor, semantic conditioning can only
improve on a strong structural prior, not silently degrade it.

We evaluate the method on three real labeled property graphs - adverse-event
reporting, financial-crime filings, and movie recommendation - spanning binary classification, multi-class prediction, regression, and learning-to-rank tasks. The router improves over the strongest
static-semantic baseline on the harder, imbalanced tasks while remaining
competitive on a balanced, near-saturated control, and the same mechanism
transfers across task types with only its prediction head changed.

\section{Related Work}

Classical message-passing GNNs aggregate information from local graph
neighborhoods. GCN uses normalized neighborhood convolution
\cite{kipf2017gcn}, GraphSAGE introduces inductive neighborhood aggregation
\cite{hamilton2017graphsage}, and GAT learns attention weights between connected
nodes \cite{velivckovic2018gat}. These methods primarily assume node-attributed
graphs and do not explicitly address target-conditioned interpretation of rich
node and relationship properties.

For heterogeneous and multi-relational graphs, relation-aware models maintain
per-relation transformations, as in R-GCN \cite{schlichtkrull2018rgcn}, and
hierarchical attention models learn importance at two levels - within a relation
and then across relations. BR-GCN \cite{iyer2024brgcn}, for example, combines
node-level attention inside each relation-specific subgraph with relation-level
attention across relations. Our routing adopts the same two-level organization,
but its attention is conditioned on a language-model query derived from property
text rather than learned from graph representations alone.

Graph Transformers extend self-attention to broader structural contexts.
Graphormer incorporates structural encodings into Transformer attention
\cite{ying2021graphormer}, while GraphGPS combines local message passing with
global attention \cite{rampavsek2022gps}. Their attention mechanisms are,
however, learned from graph representations rather than conditioned by an
adapted language model interpreting property-level evidence.

Recent studies combine graphs with language models along several lines. One line
expresses graphs for language models or uses them for graph reasoning, as in
GraphGPT \cite{tang2023graphgpt} and GraphLLM \cite{jin2023graphllm}. A second
line adapts a language model parameter-efficiently as a graph encoder or
predictor: BiGTex \cite{bigtex2025} and GaLoRA \cite{galora2026} keep the
language model frozen and use low-rank adaptation to inject structural signal for
node classification. A third, most closely related line places the language
model inside message passing on text-rich graphs: the LLM-as-graph-kernel
approach recasts the language model as the aggregation operator
\cite{zhang2026ramp}, and LEMP4HG generates language-model connection analyses
that are fused into messages through a gate \cite{lemp4hg2025}. Our method
differs from all three. We retain a standard GNN as the structural learner
rather than replacing aggregation with the language model; we use the language
model not as an encoder or predictor but as an in-propagation conditioner that
produces a target-specific routing query; and we constrain its influence to a
bounded residual on a fixed topology anchor, so semantic conditioning can refine
but not override the structural prediction. We further target labeled property
graphs, treating relationship properties and schema-level information as
first-class evidence.

\section{Method}

Let an LPG be defined as
\[
\mathcal{G}=(\mathcal{V},\mathcal{E},\tau_V,\tau_E,\mathbf{P}_V,\mathbf{P}_E),
\]
where $\mathcal{V}$ and $\mathcal{E}$ denote nodes and relationships, $\tau_V$ and $\tau_E$ are their labels or types, and $\mathbf{P}_V$ and $\mathbf{P}_E$ contain their property maps. For a target node $v$, the objective is to predict an output $y_v$ using its structural neighborhood and property-derived semantic evidence.

\subsection{Structural Anchor}

A topology GNN operates on graph connectivity and type information to produce structural node states:
\[
\mathbf{h}^{(l+1)}_v =
\operatorname{GNN}^{(l)}
\left(
\mathbf{h}^{(l)}_v,
\left\{\mathbf{h}^{(l)}_u:(u,v)\in\mathcal{E}\right\}
\right).
\]
The final structural representation $\mathbf{h}^{\mathrm{top}}_v$ is mapped to anchor logits
\[
\mathbf{z}^{\mathrm{top}}_v =
f_{\mathrm{anchor}}(\mathbf{h}^{\mathrm{top}}_v).
\]
This prediction path provides a stable reference that does not depend on the SLM successfully interpreting every property.

\subsection{Semantic LPG Encoding}

Node and relationship property maps are serialized into semantic atoms containing labels, property names, and property values. An SLM encoder maps them to fixed-dimensional representations:
\[
\mathbf{s}_v=\operatorname{Enc}(\mathbf{P}_V(v),\tau_V(v)),
\qquad
\mathbf{s}_{uv}=\operatorname{Enc}(\mathbf{P}_E(u,v),\tau_E(u,v)).
\]

For every relationship $(u,v)$ incident to target $v$, we construct a message candidate
\[
\mathbf{x}_{uv}=
\left[
\mathbf{h}^{\mathrm{top}}_u
\mathbin{\|}\mathbf{s}_u
\mathbin{\|}\mathbf{s}_{uv}
\mathbin{\|}\mathbf{e}_{\tau_E(u,v)}
\right],
\]
where $\mathbf{e}_{\tau_E(u,v)}$ represents the relationship type and $\|$ denotes concatenation. Thus, each candidate contains structural, node-semantic, relationship-semantic, and schema-level evidence.

\subsection{SLM-Conditioned Query}

Structured graph views, target semantics, and anchor decision features are projected into the embedding space of a small language model and prepended to a short textual prompt as soft tokens. The SLM is adapted using low-rank adaptation while its base parameters remain quantized. Its final hidden state produces the routing query
\[
\mathbf{q}_v=
\operatorname{norm}\left(
f_q(\operatorname{SLM}(\mathbf{T}_v))
\right),
\]
where $\mathbf{T}_v$ denotes the target-specific soft-token sequence.

The query also modulates message values using feature-wise affine transformations:
\[
\widetilde{\mathbf{v}}_{uv}
=
\mathbf{v}_{uv}\odot
\left(1+\gamma(\mathbf{q}_v)\right)
+\beta(\mathbf{q}_v).
\]
This allows the SLM to influence both message importance and message content.

\subsection{Hierarchical Relation Routing}

Routing is performed in two stages. First, candidate messages are grouped according to their relationship type $r$. Within each group, attention weights are calculated as
\[
\alpha_{uv}^{(r)}
=
\frac{
\exp\left(\mathbf{q}_v^\top \mathbf{k}_{uv}\right)
}{
\sum_{(j,v):\tau_E(j,v)=r}
\exp\left(\mathbf{q}_v^\top \mathbf{k}_{jv}\right)
}.
\]
A relation-specific summary is then obtained:
\[
\mathbf{c}_{v,r}
=
\sum_{(u,v):\tau_E(u,v)=r}
\alpha_{uv}^{(r)}\widetilde{\mathbf{v}}_{uv}
+\mathbf{e}_r.
\]
The second stage routes information across the available relation summaries:
\[
\beta_{v,r}
=
\frac{
\exp\left(\mathbf{q}_v^\top \mathbf{k}_{v,r}\right)
}{
\sum_{r'\in\mathcal{R}_v}
\exp\left(\mathbf{q}_v^\top \mathbf{k}_{v,r'}\right)
},
\]
where $\mathcal{R}_v$ is the set of relationship types incident to $v$. The final routed message is
\[
\mathbf{m}_v
=
\sum_{r\in\mathcal{R}_v}
\beta_{v,r}\mathbf{v}_{v,r}.
\]
This decomposition prevents numerous relationships of one type from directly competing with isolated relationships of another type. It also provides two complementary explanations: $\alpha_{uv}^{(r)}$ identifies relevant neighbors within a relation, while $\beta_{v,r}$ identifies relevant relation types.

\subsection{Bounded Residual Prediction}

The routed message updates the structural target state:
\[
\widetilde{\mathbf{h}}_v
=
\operatorname{LayerNorm}
\left(
\mathbf{h}^{\mathrm{top}}_v+\mathbf{m}_v
\right).
\]
A residual head predicts a bounded correction:
\[
\Delta\mathbf{z}_v
=
\tanh\left(
f_{\Delta}
\left(
[\widetilde{\mathbf{h}}_v\|\mathbf{m}_v]
\right)
\right),
\]
and the final prediction is
\[
\mathbf{z}_v
=
\mathbf{z}^{\mathrm{top}}_v+
\delta_{\max}\Delta\mathbf{z}_v.
\]
The bound $\delta_{\max}$ prevents semantic routing from arbitrarily replacing the structural prediction. For regression, the same formulation can be used with scalar anchor and residual outputs.

The model is optimized using the supervised prediction loss together with an auxiliary SLM verbalizer loss. Only the LoRA parameters, soft-token projections, routing modules, and residual head require adaptation. Cached semantic encodings and low-bit SLM quantization make training feasible on a single consumer GPU.

\section{Implementation Details}

We use Qwen2.5-1.5B-Instruct \cite{qwen2024} as the small language model.
Node and relationship property atoms are encoded with the same model using a
maximum sequence length of 128 tokens. During hierarchical routing, the SLM is
loaded using four-bit NF4 quantization and adapted with LoRA
\cite{hu2022lora,dettmers2023qlora}. LoRA is applied to the query, key, value,
and output projections with rank 8 and scaling parameter 16. The base SLM
parameters remain frozen.

The routing model uses up to 32 incident relationships per target. Training uses five-fold stratified cross-validation; results are averaged over five random seeds for all tasks. Semantic
property encodings are cached before training. Quantized SLM computation uses
bfloat16 precision, gradient clipping, and a short LoRA-only warm-up period.
These settings allow the complete model to be trained using a single NVIDIA
RTX 3090 GPU.

\subsection{Datasets}

We evaluate on three labeled property graphs from the Neo4j~\cite{neo4j} graph-examples
collection, each turned into a node-prediction task in which the label-defining
attributes are removed from the encoded features.

The \textbf{Healthcare
Analytics\footnote{\url{https://github.com/neo4j-graph-examples/healthcare-analytics}}} graph
is built from FDA Adverse Event Reporting System records, linking cases to
drugs, indications, reactions, therapies, report sources, and demographics
(11{,}948 nodes, 91{,}090 edges). The targets are 4{,}307 \texttt{Case} nodes
labeled by death outcome (7.6\% positive); outcome records and report
identifiers are used only to form labels and are excluded from the encoded
features.

The \textbf{FinCEN Files\footnote{\url{https://github.com/neo4j-graph-examples/fincen}}} graph is derived
from the ICIJ investigation of leaked Suspicious Activity Reports, connecting
\texttt{Entity}, \texttt{Filing}, and \texttt{Country} nodes via originator,
beneficiary, and concern relationships (7{,}524 nodes, 40{,}835 edges). The
targets are 4{,}507 \texttt{Filing} nodes labeled by whether the transaction
amount lies in the top quartile (25.0\% positive); the amount is removed from
the features.

The \textbf{Recommendations\footnote{\url{https://github.com/neo4j-graph-examples/recommendations}}} graph
extends the MovieLens corpus with movie metadata, connecting \texttt{Movie},
\texttt{User}, \texttt{Actor}, \texttt{Director}, and \texttt{Genre} nodes
(28{,}863 nodes, 166{,}261 edges). The targets are the 9{,}058 rated
\texttt{Movie} nodes labeled by whether the IMDb rating exceeds the median of
6.9 (50.1\% positive); rating and vote count are removed.

\subsection{Evaluation Setup}

Each task was evaluated with five stratified folds per seed; we report the mean
and standard deviation across seed-level means, using five seeds for all tasks. Because the positive class ranges from 7.6\% (healthcare) to 50.1\%
(IMDb), macro-F1 and AUROC are emphasized over accuracy, which is inflated by
the majority class on the imbalanced tasks.

The proposed model was compared with three architectural baselines:
a topology-only GNN, a node-semantic static GNN, and an edge-semantic static
GNN. The node-semantic model initializes nodes using SLM-derived property
encodings, whereas the edge-semantic model additionally incorporates encoded
relationship properties into message propagation. In both semantic baselines,
the encodings remain static and do not perform target-conditioned routing.

\section{Experimental Results}

\begin{figure*}[t]
  \centering
  \includegraphics[width=\textwidth]{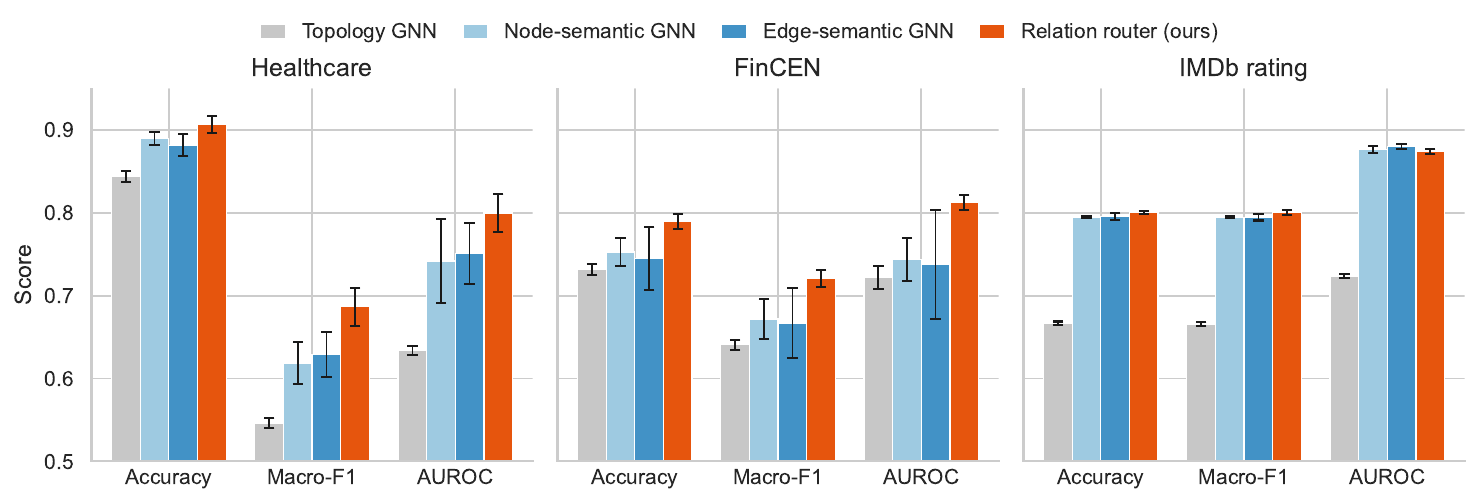}
  \caption{Binary classification on labeled property graphs (5-fold $\times$ 5-seed;
  error bars are std across seeds). The SLM-conditioned relation router (orange) is
  compared with three frozen GNN anchors.}
  \label{fig:binary_results}
\end{figure*}

On the healthcare task, the hierarchical relation router obtains the best mean
result on all three reported metrics (Table~\ref{tab:binary-router},
Fig.~\ref{fig:binary_results}). Relative to the strongest static baseline for
each metric, it improves accuracy by 1.7 points ($0.907$ vs.\ $0.890$),
macro-F1 by 5.8 points ($0.687$ vs.\ $0.629$), and AUROC by 4.8 points
($0.800$ vs.\ $0.751$). The gains in macro-F1 and AUROC are the most relevant
for this imbalanced task, where the positive class accounts for only 7.6\% of
targets and accuracy is dominated by the majority class. The small variation
across seeds (standard deviation $\leq 0.023$) indicates that the improvement
is not attributable to a single favorable data split.

The same pattern holds on FinCEN, where the router is again best on all three
metrics, improving over the strongest static baseline by 3.7 points in accuracy,
4.9 in macro-F1, and 6.9 in AUROC. On the more saturated IMDb task - where the
classes are balanced and static semantic features already capture most of the
signal - the router remains best on accuracy and macro-F1 but is marginally
surpassed by the edge-semantic baseline on AUROC ($0.874$ vs.\ $0.880$),
consistent with its status as an external-knowledge control.

\begin{table}[t]
\centering
\caption{Binary classification on labeled property graphs, 5-fold $\times$ 5-seed
(mean\,$\pm$\,std). All models share a fixed topology-GNN anchor; the relation router
predicts a bounded SLM-conditioned correction to it. Best per column per dataset in
\textbf{bold}.}
\label{tab:binary-router}
\setlength{\tabcolsep}{6pt}
\renewcommand{\arraystretch}{1.15}
\resizebox{\textwidth}{!}{%
\begin{tabular}{llcccc}
\toprule
Dataset & Model & Accuracy & Macro-F1 & AUROC & MCC \\
\midrule
\multirow{4}{*}{\shortstack[l]{\textbf{Healthcare}\\\footnotesize $n{=}4307$, 7.6\% pos.}}
 & Topology GNN           & $0.844 \pm 0.007$ & $0.546 \pm 0.006$ & $0.634 \pm 0.005$ & $0.102 \pm 0.013$ \\
 & Node-semantic GNN      & $0.890 \pm 0.008$ & $0.619 \pm 0.025$ & $0.742 \pm 0.051$ & $0.243 \pm 0.051$ \\
 & Edge-semantic GNN      & $0.882 \pm 0.013$ & $0.629 \pm 0.027$ & $0.751 \pm 0.037$ & $0.265 \pm 0.054$ \\
 & Relation router (ours) & $\mathbf{0.907 \pm 0.010}$ & $\mathbf{0.687 \pm 0.023}$ & $\mathbf{0.800 \pm 0.023}$ & $\mathbf{0.381 \pm 0.045}$ \\
\midrule
\multirow{4}{*}{\shortstack[l]{\textbf{FinCEN}\\\footnotesize $n{=}4507$, 25.0\% pos.}}
 & Topology GNN           & $0.732 \pm 0.007$ & $0.641 \pm 0.006$ & $0.722 \pm 0.014$ & $0.287 \pm 0.014$ \\
 & Node-semantic GNN      & $0.753 \pm 0.017$ & $0.672 \pm 0.024$ & $0.744 \pm 0.026$ & $0.351 \pm 0.048$ \\
 & Edge-semantic GNN      & $0.745 \pm 0.038$ & $0.667 \pm 0.042$ & $0.738 \pm 0.066$ & $0.340 \pm 0.081$ \\
 & Relation router (ours) & $\mathbf{0.790 \pm 0.009}$ & $\mathbf{0.721 \pm 0.010}$ & $\mathbf{0.813 \pm 0.009}$ & $\mathbf{0.447 \pm 0.022}$ \\
\midrule
\multirow{4}{*}{\shortstack[l]{\textbf{IMDb rating}\\\footnotesize $n{=}9058$, 50.1\% pos.}}
 & Topology GNN           & $0.667 \pm 0.002$ & $0.666 \pm 0.002$ & $0.724 \pm 0.002$ & $0.335 \pm 0.003$ \\
 & Node-semantic GNN      & $0.795 \pm 0.001$ & $0.795 \pm 0.001$ & $0.877 \pm 0.004$ & $0.591 \pm 0.003$ \\
 & Edge-semantic GNN      & $0.796 \pm 0.004$ & $0.795 \pm 0.004$ & $\mathbf{0.880 \pm 0.003}$ & $0.592 \pm 0.007$ \\
 & Relation router (ours) & $\mathbf{0.801 \pm 0.002}$ & $\mathbf{0.801 \pm 0.003}$ & $0.874 \pm 0.003$ & $\mathbf{0.602 \pm 0.005}$ \\
\bottomrule
\end{tabular}%
}
\end{table}

Across all datasets, the gap between the topology-only and static semantic
baselines confirms that node and relationship properties carry useful
predictive evidence. However, the further improvement obtained on healthcare
and FinCEN shows that encoding this evidence statically is not sufficient by
itself. Target-conditioned selection within relationship types and across
relation-level summaries provides additional value beyond static semantic
message passing, and does so while adjusting only a bounded correction to a
fixed topology anchor.

Beyond classification, the same architecture applies without modification to
regression and learning-to-rank formulations, with only the prediction head
and loss replaced. Table~\ref{tab:fincen-multitask} summarises results on three
formulations of the FinCEN filing-amount target. For regression, the router
improves Spearman correlation from $0.270$ (best anchor) to $0.483$
($+0.213$), the largest absolute gain in the study. For ranking, where
\texttt{Filing} nodes are ordered by transaction amount, it improves global
rank correlation from $0.531$ (best anchor) to $0.615$ in Spearman and
average precision from $0.532$ to $0.581$ in MAP. The same routing mechanism
thus transfers across task types with only its output head changed.

\begin{table}[t]
\centering
\caption{\textbf{Multi-task generality on a single target.} The same
SLM-conditioned relation router - with only its prediction head and loss
changed - addresses three formulations of the FinCEN filing-amount target:
binary classification (amount in the top quartile), regression (the amount
itself), and learning-to-rank (ordering filings by amount). In every
formulation it improves over the strongest GNN baseline (5-fold $\times$ 5-seed
means; metrics are macro-F1 for classification, Spearman $\rho$ and MAP for
regression and ranking). The best baseline is the top of the three GNN anchors
(topology / node-semantic / edge-semantic), which varies by formulation.}
\label{tab:fincen-multitask}
\setlength{\tabcolsep}{7pt}
\renewcommand{\arraystretch}{1.15}
\resizebox{\textwidth}{!}{%
\begin{tabular}{llccc}
\toprule
Formulation & Metric & Best GNN baseline & Router (ours) & Gain \\
\midrule
Binary (amount $\geq$ top quartile) & Macro-F1 & $0.672$ & $\mathbf{0.721}$ & $+0.049$ \\
Regression (amount)                 & Spearman & $0.270$ & $\mathbf{0.483}$ & $+0.213$ \\
Ranking (by amount)                 & Spearman & $0.531$ & $\mathbf{0.615}$ & $+0.084$ \\
Ranking (by amount)                 & MAP      & $0.532$ & $\mathbf{0.581}$ & $+0.049$ \\
\bottomrule
\end{tabular}%
}
\end{table}

Table~\ref{tab:multiclass-router} extends the evaluation to multi-class
prediction. On healthcare severity (5 classes), the router improves accuracy
by 9.1 points over the strongest static baseline ($0.555$ vs.\ $0.464$) and
AUROC by 4.0 points ($0.714$ vs.\ $0.674$), with markedly lower variance
across seeds. On FinCEN amount (4 ordinal classes), it improves accuracy by
2.4 points ($0.502$ vs.\ $0.478$) and AUROC by 2.0 points ($0.764$ vs.\
$0.744$). In both cases the router sweeps every reported metric, confirming
that target-conditioned routing generalises beyond binary tasks.

\begin{table}[t]
\centering
\caption{Multi-class classification on labeled property graphs, 5-fold $\times$ 5-seed
(mean\,$\pm$\,std). All models share a fixed topology-GNN anchor; the relation router
predicts a bounded SLM-conditioned correction to it. AUROC is one-vs-rest
macro-averaged. Best per column per dataset in \textbf{bold}.}
\label{tab:multiclass-router}
\setlength{\tabcolsep}{6pt}
\renewcommand{\arraystretch}{1.15}
\resizebox{\textwidth}{!}{%
\begin{tabular}{llcccc}
\toprule
Dataset & Model & Accuracy & Macro-F1 & AUROC & MCC \\
\midrule
\multirow{4}{*}{\shortstack[l]{\textbf{Healthcare severity}\\\footnotesize $n{=}4307$, 5 classes}}
 & Topology GNN           & $0.393 \pm 0.010$ & $0.216 \pm 0.006$ & $0.615 \pm 0.008$ & $0.109 \pm 0.014$ \\
 & Node-semantic GNN      & $0.464 \pm 0.036$ & $0.272 \pm 0.034$ & $0.674 \pm 0.050$ & $0.200 \pm 0.054$ \\
 & Edge-semantic GNN      & $0.463 \pm 0.027$ & $0.264 \pm 0.024$ & $0.670 \pm 0.029$ & $0.193 \pm 0.041$ \\
 & Relation router (ours) & $\mathbf{0.555 \pm 0.025}$ & $\mathbf{0.327 \pm 0.012}$ & $\mathbf{0.714 \pm 0.013}$ & $\mathbf{0.288 \pm 0.035}$ \\
\midrule
\multirow{4}{*}{\shortstack[l]{\textbf{FinCEN amount}\\\footnotesize $n{=}4507$, 4 classes}}
 & Topology GNN           & $0.415 \pm 0.003$ & $0.415 \pm 0.004$ & $0.692 \pm 0.003$ & $0.222 \pm 0.005$ \\
 & Node-semantic GNN      & $0.478 \pm 0.014$ & $0.482 \pm 0.016$ & $0.744 \pm 0.018$ & $0.308 \pm 0.019$ \\
 & Edge-semantic GNN      & $0.467 \pm 0.032$ & $0.468 \pm 0.041$ & $0.731 \pm 0.038$ & $0.294 \pm 0.043$ \\
 & Relation router (ours) & $\mathbf{0.502 \pm 0.004}$ & $\mathbf{0.504 \pm 0.003}$ & $\mathbf{0.764 \pm 0.003}$ & $\mathbf{0.339 \pm 0.005}$ \\
\bottomrule
\end{tabular}%
}
\end{table}

\section{Discussion}

A central design choice in our method is to keep the structural prediction and
the semantic correction separate: a frozen topology anchor fixes a strong
structural prior, and the language model may only add a bounded residual on top
of it. This separation is what makes the role of the SLM interpretable. Rather
than asking whether a language-model-driven model is better in aggregate, it
isolates the contribution of semantic conditioning as a quantity that is zero at
initialization and bounded thereafter, so any gain is attributable to the
learned routing rather than to re-selecting a stronger predictor. The same
choice makes the method deliberately conservative: it cannot trade structural
accuracy for semantic plausibility, which is desirable when the anchor is
already strong, but it also limits how far semantic evidence alone can carry a
prediction when the structural signal is weak. We regard this as an appropriate
default for property-rich graphs, where structure is usually reliable and
semantics is complementary rather than primary.

The experiments also suggest a simple principle for when such conditioning is
worthwhile. Target-conditioned routing can help only to the extent that the
evidence relevant to a prediction is selective - concentrated in particular
neighbors and relationship types rather than spread uniformly across a node's
neighborhood. When the relevant signal is diffuse, or already summarized by
target-independent pooling, a static semantic encoder is close to optimal and
there is little for routing to recover. This framing predicts that the benefit
of the architecture should track the heterogeneity and selectivity of a task's
relational evidence rather than its raw difficulty, and it offers a way to
anticipate, before training, where the method is most likely to pay off.

This positions the approach between two common alternatives. Static property
encoders make semantic information available to a GNN but fix it before
propagation, so they cannot reweight evidence with respect to the target;
graph-to-text language-model approaches gain contextual interpretation but
discard explicit structure and are constrained by the context window.
Conditioning a real message-passing process on a small language model keeps the
structural learner intact while letting semantics modulate it locally. A
practical consequence is interpretability: the within- and across-relation
weights expose, for each prediction, which neighbors and which relationship
types were selected, which is directly useful for auditing decisions in domains
such as financial-crime or clinical-outcome analysis.

\section{Conclusion}

We presented SLM-Conditioned Hierarchical Relation Routing, a method for
integrating language-derived semantics directly into message propagation over
labeled property graphs. Rather than treating node and edge properties as static
inputs, the architecture uses a parameter-efficient SLM to condition which
neighboring messages and relationship types are selected for each target. A
two-stage routing mechanism separates within-type from across-type evidence
selection, and a bounded residual update keeps the structural anchor stable
under semantic corrections.

The routing mechanism transfers across binary classification, multi-class
prediction, regression, and learning-to-rank formulations with only the
prediction head and loss changed, and improves over the strongest
static-semantic baseline in each case. The results support the view that
relational context - not just richer node features - is a meaningful source
of signal in property-rich graphs, and that an SLM is a practical means of
exploiting it during graph learning.

\begin{credits}
\subsubsection{\ackname} This manuscript acknowledges the use of Claude Code~\cite{claude}, powered by the Claude Opus 4.8, to improve language clarity, refine sentence structure, and enhance overall writing precision.
\end{credits}


%
%
%
%

\end{document}